# Expandable YOLO: 3D Object Detection from RGB-D Images*

Masahiro Takahashi[1], Alessandro Moro[2],
Yonghoon Ji[1], *Member, IEEE*, and Kazunori Umeda[1], *Member, IEEE*

*Abstract*— This paper aims at constructing a light-weight object detector that inputs a depth and a color image from a stereo camera. Specifically, by extending the network architecture of YOLOv3 to 3D in the middle, it is possible to output in the depth direction. In addition, Intersection over Uninon (IoU) in 3D space is introduced to confirm the accuracy of region extraction results. In the field of deep learning, object detectors that use distance information as input are actively studied for utilizing automated driving. However, the conventional detector has a large network structure, and the real-time property is impaired. The effectiveness of the detector constructed as described above is verified using datasets. As a result of this experiment, the proposed model is able to output 3D bounding boxes and detect people whose part of the body is hidden. Further, the processing speed of the model is 44.35 fps.

## I. Introduction

Detecting or recognizing objects and counting the number of objects is a hot topic and in demand in the field of security and marketing recently. However, some of the problems faced are human error from carelessness or labor costs if hiring people. For solving these problems, it is necessary to enhance the camera system that realizes automated object detection.

Due to the breakthrough progress of deep learning in recent years, object detectors have very high accuracy. In particular, three types of network structures have achieved important results. The first is Region-based CNN (R-CNN) [1] type represented by Fast/Faster/Mask [2, 3, 4] R-CNN. Among them, Mask R-CNN realizes instance segmentation by putting Fully Convolutional Network (FCN) [5] in the network. However, this type has a convolutional neural network for each object. For this reason, the network structure and the calculation cost are large. The second is Single Shot MultiBox Detector (SSD) [6] type. Since SSD is faster than R-CNN and reflects the feature extraction results for each scale, it has the advantage of being robust even if multiple objects exist in the scene. However, SSD has many arbitrary parameters such as the selection of scale and size setting of the basic rectangle. Finally, You Only Look Once (YOLO) [7] is composed of one simple network, and YOLOv2 [8] achieved higher accuracy and processing speed than SSD. However, due to the limitations of the algorithm, it is difficult to detect when the objects are adjacent if they are in proximity to the same anchor point (object merging).

On the other hand, in the counting of people and automatic driving, it is necessary to detect the object considering the depth information of the object in order to cope with the occlusion between objects. However, these network architectures are huge, and not real-time. Moreover, these take only point clouds as input, and color information cannot be taken into account. There are also detectors that emphasize real-time characteristics, such as Complex-YOLO [9] and YOLO3D [10]. However, these use detection methods that use point clouds from a bird's-eye view. Thus, it is necessary to use another process such as associating with a subjective image and to use a sensor that can acquire a wide range of points such as Laser Imaging Detection and Ranging (LiDAR).

Therefore, the purpose of this study is to construct an object detector that takes depth information and color information as input and identifies the 3D position of the object. In addition, we aim to construct a faster system by reducing the weight of the network. This makes it possible to easily and quickly identify the 3D position using a stereo camera, which was conventionally done by expensive range finders such as LiDAR. In the proposed model, a 3D bounding box is an output based on the structure of YOLO. This can be expected to be robust even for scenes with occlusions that are difficult to separate by simply combining 2D based methods and depth images. In chapter 2, we describe the network architecture and the loss function, and in chapter 3, we explain the verification experiment using the datasets.

## II. Expandable YOLO

### A. Network Architecture

The structure of the proposed network is shown in Fig. 1. We named it "Expandable YOLO" (hereinafter, this is called "E-YOLO"). E-YOLO is conceptually simple: YOLOv3 [11] used a color image consisting of RGB channels as input; we add a new depth image channel to this and input it as a single image. From this, the network is able to extract features from color and depth images at the same time. In E-YOLO, the image of 4after channels completed in this way is used as input.

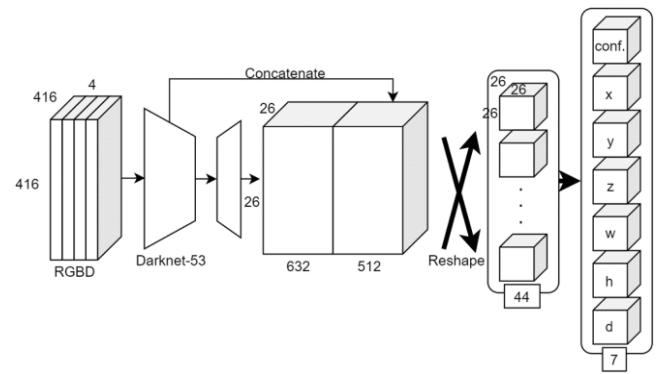

Figure 1. The structure of E-YOLO.

[1]The Course of Precision Engineering, School of Science and Engineering, Chuo University, 1-13-27 Kasuga, Bunkyo-ku, Tokyo, Japan m.takahashi@sensor.mech.chuo-u.ac.jp.
[2]RITECS Inc., 3-5-11 Shibasaki, Tachikawa-shi, Tokyo, Japan.

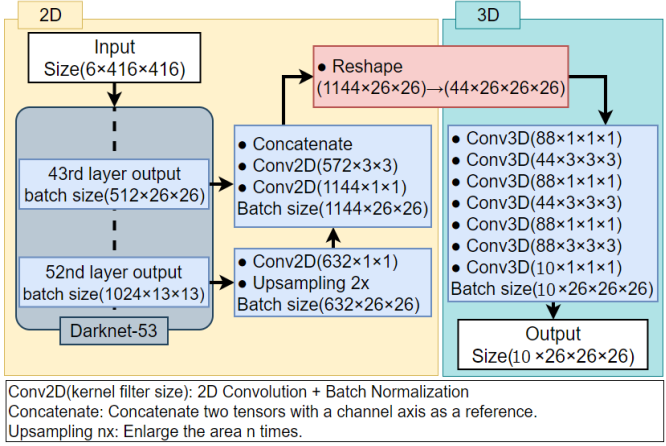

Figure 2. The parameters of E-YOLO

Darknet-53, which is used in YOLOv3, is adopted as the network structure for extracting features up to the middle stage. The reason is that in YOLOv3, the feature extraction for predicting the bounding box was successful, and compared with the case where it was performed by Residual Network (ResNet) [12], the score was high in accuracy and processing speed. Therefore, Darknet-53 is an efficient network structure when extracting features from 2D images. In E-YOLO, the depth information is included in the input, but since this is input as a depth image, we think that feature extraction can be performed as in the case of a color image.

Next, we explain the part of the network structure that outputs a 3D bounding box from the features using Darknet-53. In this part, two outputs are obtained at different scales as shown in Fig. 2. These two layers are connected using upsampling, and convolution with 1×1 kernel filter is performed. After that, it is divided into 26 channels in the channel direction. This makes the tensor 3D square. We inspired in this part by unified detection, a concept common to YOLO. Unified detection is a concept that allows classification and region identification to be output at the same time by storing the bounding box coordinates and classification results for each channel. Based on this concept, if the type of output value could be controlled for each channel, the same could be done for each channel range. Here, Feature Pyramid Network (FPN) [13] structure is adopted in YOLOv3; however, the proposed model handles only the scale of the middle part to reduce the calculation cost.

With this network architecture, E-YOLO outputs a 3D bounding box from the color image and the depth image. All convolutional layers use 3×3 or 1×1 kernel filter as shown in Fig. 2. In addition, the size of each channel of the input image is 416×416 and the output size is 26×26×26 because the shape is transposed from 2D to 3D. Therefore, E-YOLO can detect the scene in Fig. 3. The number of input channels is 4; R, G, B of the color image and depth. On the other hand, the number of output channels is 10. Of these channels, the first 8 channels are for outputting 3D bounding boxes, which are confidence value, unreliable value, bounding box's anchor point x, y, and z, the size of bounding box width, height, and depth. And the others are for recognizing the classes (person or object). We set the number of channels for classification to the same value of YOLOv3's one. In preliminary experiments, we estimated that the best performances were obtained by setting these kernel sizes.

### B. Implementation Details

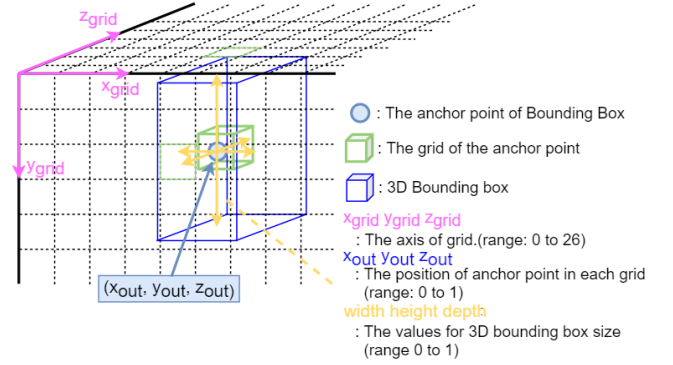

Figure 3. The model output details.

We set hyper-parameters of Darknet-53 following existing YOLOv3 work. The other parameters were set as shown in Fig. 2. But to reduce the calculation cost, the number of bounding boxes B in each grid is set to 1

The training method is described below. Compared to YOLOv3, the proposed model also output in the depth direction. Therefore, we use the loss function of the YOLO3D loss function including the square error for depth information for training. The loss function is as follows:

$$L_{ExpandableYOLO} = \lambda_{coord} \sum_{i=1}^{G} l_i^{obj} \left[ \left(t_x^{(i)} - \hat{t}_x^{(i)}\right)^2 + \left(t_y^{(i)} - \hat{t}_y^{(i)}\right)^2 + \left(t_z^{(i)} - \hat{t}_z^{(i)}\right)^2 \right]$$
$$+ \lambda_{coord} \sum_{i=1}^{G} l_i^{obj} \left[ \left(t_w^{(i)} - \hat{t}_w^{(i)}\right)^2 + \left(t_h^{(i)} - \hat{t}_h^{(i)}\right)^2 + \left(t_d^{(i)} - \hat{t}_d^{(i)}\right)^2 \right]$$
$$+ \lambda_{coord} \sum_{i=1}^{G} l_i^{obj} \left(c_{obj}^{(i)} - \widehat{c_{obj}}^{(i)}\right)^2 + \lambda_{noobj} \sum_{i=1}^{G} l_i^{obj} \left(c_{obj}^{(i)} - \widehat{c_{obj}}^{(i)}\right)^2$$
$$+ \sum_{i=1}^{G} \sum_{k=1}^{K} l_i^{obj} \left(p_k^{(i)} - \widehat{p_k}^{(i)}\right)^2 \quad (1)$$

Almost all values in equation (1) have the same meanings as YOLO3D's ones. The value of G = (Height/26) × (Width/26) × (Depth/26) is the number of cells in the grid. In addition, λ, which indicates the usage of the error that is arbitrarily set, is set to 1 for the error of the center coordinate and size of the bounding box, and to 10 for the error when no object exists. The variable $t$ is each value that composes the bounding box (x, y, z of the anchor point, and the width and height), and $c$ represents the confidence that the bounding box exists at the position. The symbol p is defined as a variable indicating the probability of belonging to a class. The identifier l indicates whether or not the bounding box included in the supervised data exists in the cell. If the bounding box exists, errors with respect to the bounding box and classification are calculated, and if not, the reliability value portion of the object is calculated.

The selection of the optimal bounding box candidates is described as follows. A typical method is a Non-Maximum Suppression (NMS). This is a method of eliminating the bounding box estimated for the same object based on the score representing the degree of overlap of the region called Intersection over Union (IoU). Here, conventional object detectors that handle 3D information such as YOLO3D calculate IoU for 2D from the two directions of the sensor front and vertical directions, and select the bounding box by NMS. However, with this method, IoU is calculated twice for the same 3D bounding box, which is inefficient when performing parallel calculations on the GPU. Therefore, in the proposed model, 3D IoU representing the degree of volume overlap as shown in Fig. 4 is defined, and NMS is executed based on this. As a result, the IoU calculation is performed only once on the GPU, and an improvement in processing speed can be expected. As for the threshold of IoU in NMS, 0.5 was used in YOLO, R-CNN, and SSD. But it is better to use 0.35 in the case of volume ratio.

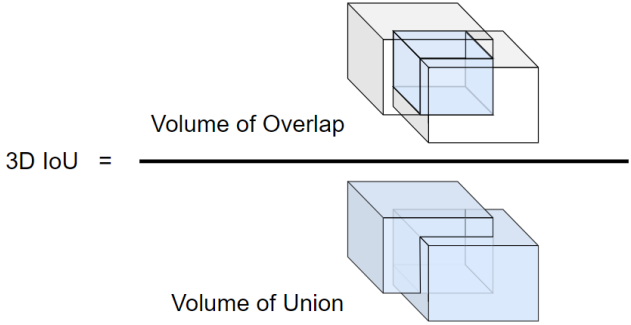

Figure 4. The 3D IoU for NMS

## III. EXPERIMENTS

In order to confirm the effectiveness of the proposed model, the accuracy of the model was verified using multiple datasets. In addition, the processing time was measured for each part of the implemented system in order to verify the real-time property.

### A. Training with Original Dataset

We trained the model using the dataset we created, measured the time required for learning, and evaluated the model accuracy using the IoU score. We acquired data using RealSense D435 [14] in room 2720 on the Korakuen Campus of Chuo University. As shown in Fig 5, the data was taken indoors as people were walking. The label was created automatically by Mask R-CNN, and manually labeled those with a lower confidence value. Labeling of the depth value was performed using clustering for point clouds generated from RGB-D. The depth was normalized to a range of 10 meters. We trained the model using the 1240 scene data obtained in this way. This data was divided into 1140 pieces of training data and 100 pieces of validation data. we set the learning rate to 0.001 and selected Adaptive moment estimation (Adam) [15] as the optimization method. The machine spec at the time of training was GPU: NVIDIA RTX 2080 and CPU: Intel Core i7 8700K. The training was done end-to-end including Darknet.

The transition of the loss indicating the learning progress when learning 100 epochs is as shown in Fig. 6. The vertical axis represents the loss, the horizontal axis represents the number of epochs, and the loss value represents the average loss in one epoch. The minimum loss value for training data was 2337.63, and the minimum loss value for validation data was 7180.24 (Table I).

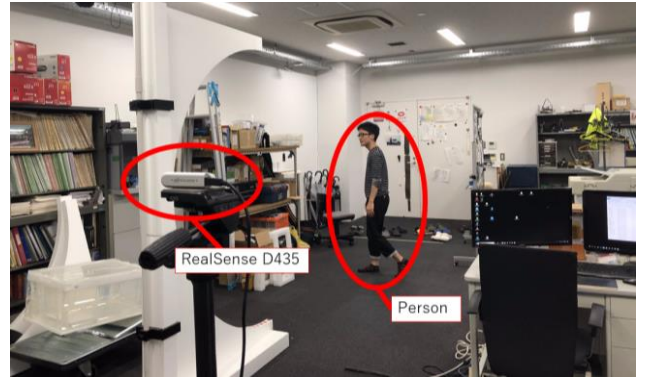

Figure 5. Environment when collecting data.

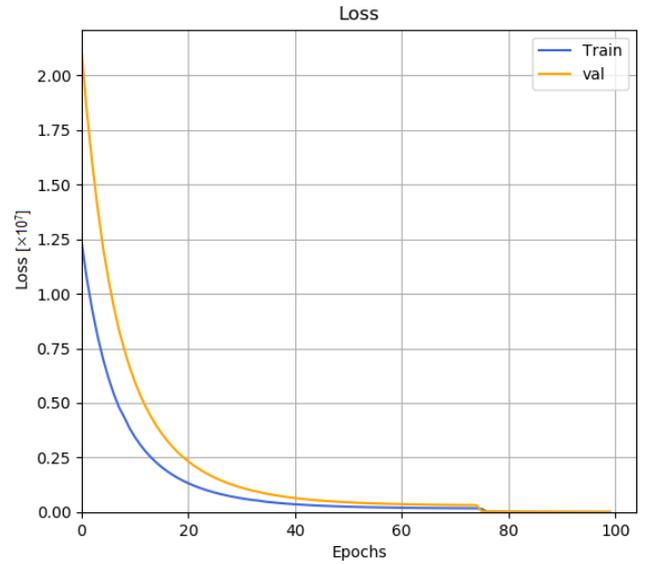

Figure 6. The transition of the loss.

TABLE I. THE MINIMUM LOSS VALUE

| Data Type | Minimum Loss Value |
|---|---|
| Training | 2337.63 |
| Validation | 7180.24 |

Although the minimum loss value looks large, from the loss transition, the loss with respect to the training data is less than 1/5000 of the initial loss. Looking at the transition of loss for validation data, the proposed model is robust even for unlearned data. In addition, we believe that errors can be further reduced by changing the feature extractor from Darknet to other networks. It is because Darknet is optimized for YOLOv3 architecture, not for 3D convolutional networks. So it is needed to optimize the parameters of network with any automatic machine learning method [16].

## B. Verification of Detection Accuracy

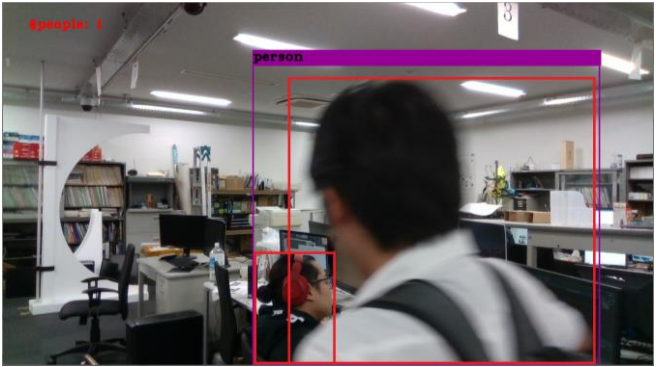

Figure 7. YOLOv3 detection result. In this scene, there are two people (The red bounding boxes). But YOLOv3 collectively outputs one person (The purple bounding box).

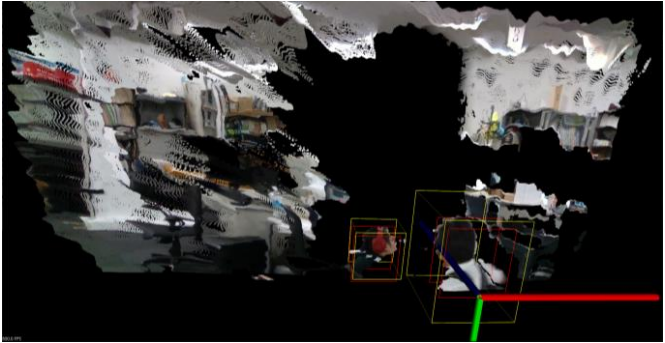

Figure 8. E-YOLO detection result. In this scene, The ground truths are the red bounding boxes. And The outputs of E-YOLO are yellow bounding boxes. The outputs are able to devide two people although a little big.

We verified the proposed model of detection accuracy. Fig. 7 and 8 show the detection results of the trained models of YOLOv3 and E-YOLO. The detection results using the proposed model were rendered using the Point Cloud Library (PCL) [17]. In these figures, the red rectangle represents the ground truth. The bounding box color of the detection result is shown in purple in Fig. 7 and yellow in Fig. 8. Comparing these results, it can be seen that the proposed model can detect two-dimensional objects in the same way as YOLOv3, and can also estimate the depth direction at the same time. In addition, while the original YOLO did not detect the central person wearing black clothes, our proposed model could detect this. However, the accuracy of the length in the depth direction for bounding boxes was still insufficient compared with the ground truth.

In order to quantitatively evaluate the accuracy of the estimated bounding box, we used the IoU score. We also evaluated it with the 3D IoU described in Fig. 4 because the proposed model outputs a 3D bounding box.

The results of the IoU score with the training data are shown in Table II. According to the previous studies, the YOLOv3 IoU score is about 0.7 to 0.8 for every class. Thus, the accuracy of the bounding box in the 2D dimension of the proposed model has decreased slightly. This is because the proposed model did not adopt the FPN structure used in YOLOv3. By using the FPN structure, the robustness for objects of different sizes is improved, but the processing speed is reduced. Therefore, it is expected to improve the accuracy while maintaining the real-time property by partially incorporating it into the network in consideration of the processing speed. We also compared 2D IoU and 3D IoU scores. At first glance, the 3D IoU score appears to be low; however, 2D IoU corresponds to 3D IoU$^{2/3}$ from the relationship between the volume ratio and the area ratio. From this result, it is considered that the proposed model can extract 2D and 3D features to the same extent. Therefore, in order to improve accuracy, improvements such as multi-layering are required in the part of the network architecture that performs both 2D convolution and 3D convolution.

TABLE II. THE IoU SCORES OF PROPOSED METHOD

|  | 2D IoU | 3D IoU | 3D IoU$^{2/3}$ |
|---|---|---|---|
| Mean | 0.54 | 0.39 | 0.53 |
| Max | 0.92 | 0.85 | 0.90 |

## C. Verification of Processing Speed

We verified the processing speed of the proposed model. The machine spec used for verification was GTX 1080 Ti for GPU and Core i7 4790 for CPU. The model was implemented with PyTorch [18]. The results are shown in Table III. As a result, although the processing speed of the proposed model was lower than that of YOLOv3, it was able to operate at a very high speed as a network to obtain the 3D output. This is because the number of 3D convolution layers is reduced by using 2D convolution layers for reature extracting. However, since it is necessary to increase the number of layers in order to improve the accuracy of object detection, by improving the efficiency of the RGB-D feature extraction method, the accuracy is improved while maintaining the processing speed.

TABLE III. THE IoU SCORES OF PROPOSED METHOD

| Method | GPU | SPEED [fps] |
|---|---|---|
| YOLOv3 [11, 19] | GTX 1080Ti | 74 |
| Complex-YOLO [9] | Titan X | 50.4 |
| E-YOLO | GTX 1080Ti | 44.4 |
| YOLO3D [10] | Titan X | 40 |
| VoxelNet [20] | Titan X | 4.3 |

## IV. CONCLUSION

In this study, we constructed a lightweight network that can output a 3D bounding box from RGB-D images. By using the depth image instead of the point cloud to acquire the features in the depth direction by 2D convolution, it is possible to realize a lighter network than 3D processing. In our future work, we aim to extend the proposed network to perform instance segmentation.

## REFERENCES

[1] R. Girshick, J. Donahue, T. Darrell, and J. Malik, "Rich feature hierarchies for accurate object detection and semantic segmentation," in *Proc. of the IEEE Conf. on Computer Vision and Pattern Recognition (CVPR)*, pp. 580-587, 2014.

[2] R. Girshick, "Fast R-CNN," in *Proc. of the IEEE Int. Conf. on Computer Vision (ICCV)*, pp. 1440-1448, 2015.

[3] S. Ren, K. He, R. Girshick, and J. Sun, "Faster R-CNN: Towards real-time object detection with region proposal networks," Conference on Neural Information Processing Systems (NIPS), 2015.
[4] K. He, G. Gkioxari, P. Dollar, and R. Girshick, "Mask R-CNN," in *arXiv preprint arXiv:1703.06870*, 2017.
[5] J. Long, E. Shelhamer, and T. Darrell, "Fully Convolutional Neural Networks for Semantic Segmentation," in *Proc. of the IEEE Conf. on Computer Vision and Pattern Recognition (CVPR)*, pp. 3431-3440, 2015.
[6] W. Liu, D. Anguelov, D. Erhan, C. Szegedy, S. Reed, C. Y. Fu, and A. C. Berg, in *arXiv preprint arXiv:1512.02325*, 2015.
[7] J. Redmon, S. Divvala, R. Girshick, and A. Farhadi, "You Only Look Once: Unified, Real-Time Object Detection," in *Proc. of the IEEE Conf. on Computer Vision and Pattern Recognition (CVPR)*, pp. 779-788, 2016.
[8] J. Redmon and A. Farhadi, "YOLO9000: Better, Faster, Stronger," in *Proc. of the IEEE Conf. on Computer Vision and Pattern Recognition (CVPR)*, 2017.
[9] M. Simon, S. Milz, K. Amende, and H. M. Gross, "Complex-YOLO: An Euler-Region-Proposal for Real-Time 3D Object Detection on Point Clouds," in Proc. of the European Conference on Computer Vision (ECCV), pp. 197-209, 2018.
[10] W. Ali, S. Abdelkarim, M. Zahran, M. Zidan, and A. E. Sallab, "YOLO3D: End-to-end real-time 3D Oriented Object Bounding Box Detection from LiDAR Point Cloud," *arXiv preprint arXiv:1808.02350*, 2018.
[11] J. Redmon and A. Farhadi, "YOLOv3: An Incremental Improvement," in *arXiv preprint arXiv:1804.02767*, 2018.
[12] K. Simonyan and A. Zisserman, "Very Deep Convolutional Networks for Large-Scale Image Recognition," in *arXiv preprint arXiv:1409.1556*, 2014.
[13] T. Y. Lin, P. Dollar, R. Girshick, K. He, B. Hariharan, and S. Belongie, "Feature Pyramid Networks for Object Detection," in *Proc. of the IEEE Conf. on Computer Vision and Pattern Recognition (CVPR)*, pp. 2117-2125, 2017.
[14] Intel® RealSense™ Depth Camera D435, https://www.intelrealsense.com/depth-camera-d435/, 2020/01/30.
[15] D. P. Kingma and J. Ba, "Adam: A Method for Stochastic Optimization," in *Proc. of the 3$^{rd}$ Int. Conf. on Learning Representations (ICLR)*, 2014.
[16] B. Zoph and Q. V. Lee, "Neural Architecture Search with Reinforcement Learning," arXiv preprint arXiv:1611.01578, 2016.
[17] Point Cloud Library, http://www.pointclouds.org/, 2020/01/30.
[18] PyTorch, https://pytorch.org/, 2020/01/30.
[19] PyTorch-YOLOv3, https://github.com/eriklindernoren/PyTorch-YOLOv3, 2020/01/30.
[20] Y. Zhou and O. Tuzel, "VoxelNet: End-to-End Learning for Point Cloud Based 3D Object Detection," in *Proc. of the IEEE Conf. on Computer Vision and Pattern Recognition (CVPR)*, 2018.